\documentclass[manuscript,screen]{acmart}
\AtBeginDocument{%
  \providecommand\BibTeX{{%
    \normalfont B\kern-0.5em{\scshape i\kern-0.25em b}\kern-0.8em\TeX}}}

\acmDOI{}

\acmConference[Manuscript submitted to ACM]{}{March}{2023}
\begin{document}

\title{Voice-Based Conversational Agents and Knowledge Graphs for Improving News Search in Assisted Living}

\author{Phillip Schneider}
\email{phillip.schneider@tum.de}
\orcid{0000-0001-9492-2927}
\affiliation{%
  \institution{Department of Computer Science, Technical University of Munich}
  \city{Munich}
  \country{Germany}
}

\author{Nils Rehtanz}
\email{nils.rehtanz@tum.de}
\affiliation{%
  \institution{Department of Computer Science, Technical University of Munich}
  \city{Munich}
  \country{Germany}
}

\author{Kristiina Jokinen}
\email{kristiina.jokinen@aist.go.jp}
\orcid{0000-0003-1229-239X}
\affiliation{%
  \institution{AIRC, National Institute of Advanced Industrial Science and Technology}
  \city{Tokyo}
  \country{Japan}
}

\author{Florian Matthes}
\email{matthes@tum.de}
\orcid{0000-0002-6667-5452}
\affiliation{%
  \institution{Department of Computer Science, Technical University of Munich}
  \city{Munich}
  \country{Germany}
}


\begin{abstract}
As the healthcare sector is facing major challenges, such as aging populations, staff shortages, and common chronic diseases, delivering high-quality care to individuals has become very difficult. Conversational agents have shown to be a promising technology to alleviate some of these issues. In the form of digital health assistants, they have the potential to improve the everyday life of the elderly and chronically ill people. This includes, for example, medication reminders, routine checks, or social chit-chat. In addition, conversational agents can satisfy the fundamental need of having access to information about daily news or local events, which enables individuals to stay informed and connected with the world around them. However, finding relevant news sources and navigating the plethora of news articles available online can be overwhelming, particularly for those who may have limited technological literacy or health-related impairments. To address this challenge, we propose an innovative solution that combines knowledge graphs and conversational agents for news search in assisted living. By leveraging graph databases to semantically structure news data and implementing an intuitive voice-based interface, our system can help care-dependent people to easily discover relevant news articles and give personalized recommendations. We explain our design choices, provide a system architecture, share insights of an initial user test, and give an outlook on planned future work.
\end{abstract}




\maketitle

\section{Introduction}
Digital technologies have had a profound impact on transforming the healthcare industry. The progress in hardware technology and artificial intelligence paved the way for developing sophisticated medical systems that allow for improved patient care. New technologies such as wearable smart devices, telemedicine, or electronic health records can help to address current challenges like aging populations, widespread diseases, and staff shortages. One emerging trend in the healthcare sector, and also many other industries, is the rapid adoption of conversational agents, which are intelligent computer systems with the capability to understand and generate human language \cite{mctear2016conversational}. These agents can take up different roles in assisting patients, caregivers, and doctors. Some example tasks are performing routine checks, setting up medication reminders, documenting health states, or even engaging in social small-talk \cite{valizadeh2022ai}. Aside from the aforementioned functionalities, conversational agents can also help with seeking information about a topic, which is a basic need in our everyday lives.

Having access to information sources like the daily news is especially important for the elderly and chronically ill. Individuals with severe illnesses may experience physical impairments and decreased mobility, which puts them at risk of social isolation and disconnection from the world around them. Previous research has shown that loneliness is common in older people, and it is associated with a number of adverse mental and physical health consequences \cite{luanaigh2008loneliness,crewdson2016effect}. The ability to access news information allows them to stay informed and connected to current events, and local communities, as well as maintain a sense of engagement in society in general. Even though, in the internet age, there is a virtually unlimited number of online articles available, finding relevant as well as reliable news sources can be overwhelming. For people dependent on assisted living, there are several hurdles, including a lack of technological literacy as well as cognitive, vision, or other physical impairments. In a systematic analysis of the commercial voice assistants Amazon Alexa \footnote{Amazon Alexa: https://developer.amazon.com/en-US/alexa}, Google Assistant \footnote{Google Assistant: https://assistant.google.com/}, and Siri from Apple \footnote{Apple Siri: https://www.apple.com/siri/}, we found out that they offer only a very rudimentary news search functionality, which neglects the special needs of this user group. Concerning this problem, the following research question arises: 
\begin{quote}\emph{What are possible approaches to develop conversational agents for assisted living that support an intuitive dialogue-based interaction for news search and exploration?}\end{quote}

In our study, we propose an innovative solution for conversational news search that combines knowledge graphs and voice-based natural language interfaces. We first construct a knowledge graph from news articles by organizing a corpus of text documents in a network of entities connected with semantic relations. This makes it possible to perform entity-based search queries and navigate between related topics. Then, we build a voice-based conversational agent that can access the knowledge graph stored in a graph database. Users can talk to the agent in order to search articles and receive recommendations based on their personal interests. The contributions of our study are twofold:
\begin{enumerate}
    \item We conduct an experimental analysis of the news search capabilities of three commercial voice assistants and outline their limitations.
    \item We propose an architecture of a knowledge-enhanced conversational agent for voice-based news search.
\end{enumerate}
 We hope that the insights from this research paper can advance the understanding of how people in assisted living contexts can benefit from search-centered digital assistants, particularly in the everyday example of seeking out news information. By pointing out the shortcomings of commercial voice assistants, we want to raise awareness about their limitations when it comes to user-friendly information search and discuss interaction patterns that are more accessible and intuitive. The proposed conversational agent architecture that leverages the semantic expressiveness of knowledge graphs should inspire future work on conversational search systems in healthcare and other domains.  

The remainder of this paper is structured as follows: In Section~\ref{sec:related-work}, we briefly summarize related work and emphasize the novelty of our own study. Section~\ref{sec:interaction-patterns} presents the limitations of commercial voice assistants and discusses an improved entity-based conversational search interaction. We explain the individual components of our proposed system architecture in Section~\ref{sec:architecture-overview}. We summarize preliminary insights from an initial user test in Section~\ref{sec:user-study}. Finally, Section~\ref{sec:conclusion} concludes with an outlook on future work.   

\section{Related Work}
\label{sec:related-work}
Many conversational agents are connected to a knowledge base which provides the system with an organized information collection. The latter can be used to understand and respond to user requests. It also serves as the foundation to automatically perform search tasks. Knowledge graphs are a type of knowledge base that represent this information in a graph with entities and their semantic relationships. In recent years, knowledge graphs have shown to be a powerful representation for a number of natural language processing tasks \cite{schneider2022decade}. Previous research work on combining knowledge graphs with conversational interfaces has demonstrated improvements in utterance understanding, dialogue management, and response generation. 

Considering the problem of accurately understanding a user's intent based on a short utterance, \citet{zhou2020improving} leverage both word-oriented and concept-oriented knowledge graphs to enrich conversation data with more context. The authors construct a novel dialogue system architecture which aligns semantic representations of words and entities in order to better capture individual preferences. Through extensive experiments, the authors showed that their knowledge-enhanced system yields better performance in recommendation and conversation tasks. Another study from \citet{xu2019end} proposes a dialogue system for automatic medical diagnosis. The dialogue management relies on a medical knowledge graph for topic transitions, which makes it possible to learn to request symptoms leading to a final diagnosis. The authors use both automatic and human evaluation to assess the system's performance. In the area of dialogue response generation, knowledge graphs have been applied in numerous studies as well. Most methods utilize structured information in graphs as a grounding mechanism for the responses generated by large language models that are based on the transformer architecture \cite{chen2019kbrd,zhang2020grounded,chaudhuri2021grounding}.        

Furthermore, two studies conducted by \citet{wilcock2022conversational} and \citet{vizcarra2022knowledge} are similar to our work because they also aim to implement conversational agents for accessing semantically-rich knowledge graphs in different domains. The second study also targets the user group of elderly people and discusses ways to provide health-related assistance. However, unlike our work, its focus lies on coaching interventions instead of focusing on conversational search behavior, which motivates the study at hand \cite{trippas2018informing,schneider2023investigating}. According to our analysis of related literature on conversational search systems, we are the first to implement a system that constructs a knowledge graph from German news articles and integrates it with a voice-based search agent. This new approach is particularly effective for individuals in assisted living contexts, as it enhances accessibility and ease of use.

\section{Interaction Patterns for Voice-Based News Search and Exploration}
\label{sec:interaction-patterns}
\subsection{Experimental Analysis of Commercial Voice Assistants}
Existing voice assistants, such as Amazon Alexa, Google Assistant, and Siri from Apple, offer news search as part of their functionalities. With roughly 80 million monthly active users each, these three voice assistants are the most popular and market leaders \cite{lebow2022voice}. To get an overview of this area, we relied on a large user study from \citet{newman2018future} conducted in 2018 and we systematically tested all three assistants using a predefined question catalog in order to compare their capabilities. During this experimental assessment, we found that these commercial assistants are especially limited regarding exploratory news search and often lack depth of content.

In brief, \citet{newman2018future} investigated the behavior of users when interacting with smart speakers regarding news search with a focus on Germany, the USA, and Great Britain. He discovered that most users consume news on a daily basis through short updates via command and control. In doing so, most users feel overwhelmed by the technology and the amount of news. Therefore, they appreciate that voice-based interfaces provide precise information in an easy-to-understand format. In addition, people feel that navigating the news with their voices gives them more control over the device. It has been discovered that while utilizing voice-based agents, consumers listen to the news in the morning but wish they were shorter, just up to a minute long, and more up to date. Due to having an insufficient level of control over the search process, most users complained that the news briefings were too long. In addition, the news updates were often too detailed and were recorded with very poor audio quality. Therefore, the lack of user-friendliness decreases the overall satisfaction level. Moreover, another impediment is that the news updates are frequently repeated when using different news providers. This problem can be tackled by introducing the option to skip, repeat, or select individual articles. Furthermore, the aforementioned study identified that users would like to receive more detailed information about specific topics they tried to search for. Another observed behavior is that most users only consume well-known and leading news sources. In addition, many users have expressed a desire to compile their own news briefings with news resorts they are interested in. After reading an article, a user may want to listen to other similar articles. \citet{newman2018future} also found that many users want to ask more detailed and complex questions about an event or entity from an article. In summary, topic-specific search, personalized news content, and more navigation control could significantly improve the overall user experience.

Based on our systematic comparison of the three aforementioned assistants conducted in December 2022, it became clear that a lot of room for improvement exists regarding voice-based news search, although the status quo offers already many features. To compare Siri, Alexa, and Google Assistant, a list of questions with typical user statements in the area of news search was defined. These typical user statements pertain to broad news search, where the user receives an overview, news from a specific location, news about a certain entity, as well as various control instructions to better navigate the news output. To test Siri, an iPhone XR with IOS 16.0 was used, to test Google Assistant, the iPhone app ''Assistant'' with version 1.9.64101 was used, and for Amazon Alexa, the fourth generation Amazon Alexa Echo Dot was used. The main findings revealed that the functionalities and problems of the three assistants are quite similar and they coincide with the results from \citet{newman2018future}. 

Table~\ref{tab:assistant-test} summarizes the experimental results for each examined voice assistant. For the prompt ''Tell me the news.'', all three agents only return a web search result page for the user query. This means that users need to know exactly which keywords, such as ''play'', to use in their query in order to receive a spoken response. However, even if the correct terms are used in a request, the systems often play podcasts, some of which are longer than 10 minutes. The basic control commands all work relatively well, but they are confined to playing the next podcast, repeating, or pausing the current podcast. Siri and Google Assistant showed some difficulties with skipping or repeating a podcast. In addition, many of the suggested podcasts have similar content and thus users are confronted with a lot of repetitive information. After a user has finished listening to a news podcast, the tested voice assistants do not suggest related news on the same or similar topics. 

Another observation is that these systems do not support entity-based news search, meaning that users cannot ask for news about a specific subject, such as Donald Trump or the world cup, as shown in Table~\ref{tab:assistant-test}. In the latter situation, the tested voice assistants respond with facts, web search result pages, or podcasts instead of retrieving news articles. Because these systems have no knowledge representation of current news topics, they have problems in finding local news and news about specific events. However, entity-based search is especially important for people dependent on assisted living because it simplifies the search process and can reduce the cognitive effort to find relevant news articles. Elderly and chronically ill individuals might experience a decline in memory and attention span, making it challenging to navigate complex search interfaces and identify information they need. Therefore, it is important to have a supportive conversational agent that focuses exclusively on news, since it can respond much more precisely to their needs.

\begin{table}[h]
  \caption{Summary of experimental test results for each voice assistant}
  \label{tab:assistant-test}
  \begin{tabular}{lp{3cm}p{3cm}p{3cm}p{3cm}}
    \toprule
    Focus & Search Query & Siri & Alexa & Google Assistant \\
    \midrule
    General &  &  &  & \\
    & ''Tell me the news.'' & Displays web search result page for the query. & Plays the ''100-seconds Tagesschau'' podcast. & Plays the ''100-seconds Tagesschau'' podcast. \\
    & ''Play the news.'' & Plays the ''100-seconds Tagesschau'' podcast. & Plays the ''100-seconds Tagesschau'' podcast. & Plays the ''100-seconds Tagesschau'' podcast. \\
    \midrule
    Resort &  &  &  & \\
    & ''Tell me the news about sports.'' & Plays the ''kicker news'' podcast (1-2 minutes). & Gives an update about the biggest sports clubs. & Plays the ''100-seconds Sportschau'' podcast. \\
    & ''Tell me the news about politics.'' & Cannot retrieve any news. & Plays the ''100-seconds Tagesschau'' podcast. & Plays podcast about Kosovo and Serbia conflict (13 minutes). \\
    & ''Tell me the local news.'' & Plays the ''Deutschlandfunk'' podcast (10 minutes). & Plays the ''100-seconds Tagesschau'' podcast. & Cannot retrieve any news. \\
    \midrule
    Topic &  &  &  & \\
    & ''Tell me something about the Ukraine war.'' & Plays podcast about the Ukraine war (4 minutes). & Tells one fact about the Ukraine war. & Displays web search result page for the query. \\
    & ''Play the news about the world cup.'' & Displays web search result page for the query. & Tells some general facts about the world cup (30 seconds). & Displays web search result page for the query. \\
    & ''Play the news about Donald Trump.'' & Plays podcast that contains the name Donald Trump. & Tells a summary about Donald Trump. & Displays web search result page for the query. \\
    \midrule
    Control &  &  &  & \\
    & ''Next.'' & Cannot skip podcast. & Plays next podcast.  & Plays next podcast. \\
    & ''Again.'' & Repeats podcast. & Repeats podcast. & Cannot repeat podcast. \\
    & ''Pause.'' & Pauses podcast. & Pauses podcast. & Pauses podcast. \\
    & ''Play.'' & Plays podcast. & Plays podcast. & Plays podcast. \\
  \bottomrule
\end{tabular}
\end{table}

\subsection{Entity-Based Search of News Articles}
\label{sec:entity-search}
To address the issues outlined in the previous section, we argue that an entity-based search is a useful strategy. Entity-based search can be applied to retrieve articles about concepts and things in the objective world. The article ''Would you be interested in the news? Examining Voice-Based Recommendations for Conversational News'' written by \citet{sahijwani2020would} compares different ways how to recommend news. This research paper deals with how to provide news articles to users in a voice-based conversational setting. Five alternative approaches were demonstrated in the study. The researchers look at general news briefings, news about a specific subject pertinent to a current discussion, news about a subject from a previous discussion, news on a trending subject, and a suggestion to discuss news in general. Acceptance was up to 100\% greater for entity-based suggestions compared to trending news and was 29\% higher for entity-based recommendations compared to a generic news briefing. The paper highlights that in order to keep the user engaged, it is critical to offer options that relate to concrete entities of interest. 

The information about the relevant news and the entities they include must be stored appropriately to support entity-based search. As a data structure, a knowledge graph is the most suitable choice for this task since it can handle semantic queries, which are used frequently in entity searches. In general, a knowledge graph is a knowledge representation composed of relationships and nodes. These are stored as semantic data triples. A node can have a specific number of properties and can be given different labels. A property can be seen as a key-value pair that is maintained alongside relationships or nodes. One or more properties may be present in a node or relationship. A relationship is a directed edge between two nodes having a start node and an end node that are semantically related. 

The knowledge graph's ability to map all relevant information about an article with its entities and relationships to other similar articles is a crucial factor that contributed to the decision to use it for the agent. A graph data structure is required for this, as opposed to, for example, a traditional SQL database, which is less flexible and has the drawback of requiring numerous costly joins to execute complex queries. This approach benefits the user as it allows for the recommendation of associated articles that are similar to the one they are reading. If the topic piques their interest, they may also read articles about the same person, the same country, or other entities mentioned in the article. Thus, using tags and entities in the chosen data structure enables smooth navigation and coherent topic transitions. Another advantage is the ability to quickly filter out irrelevant news articles by constraints in the graph. Given that the dynamic entity-based search can quickly retrieve news on relevant topics, the news search process is especially suitable for people with impairments regarding attention span and memory. 

Let us consider the following example of how to find a desired news article through entity-based search. An elderly person has heard that a new restaurant will soon open in her village. She wants to find out if there is any local news but fails to remember the details about this event. With our proposed system, the person can easily navigate to existing news articles by using related keywords and entities, such as restaurant, opening, the name of the village, or food, without having to know any specifics about the restaurant, like its exact name, the cuisine, or where it is located. This saves cognitive effort because the person does not need to read through an entire newspaper or listen to a long podcast in order to identify the desired information. The example demonstrates how intuitive the associative search based on entities can be, particularly to older people who might be interested mostly in specific news within their local communities and not every update about international politics or the global stock market.

\section{Overview of System Architecture}
\label{sec:architecture-overview}
\subsection{News Knowledge Graph}
\begin{figure}[ht]
  \centering
  \includegraphics[width=0.75\linewidth]{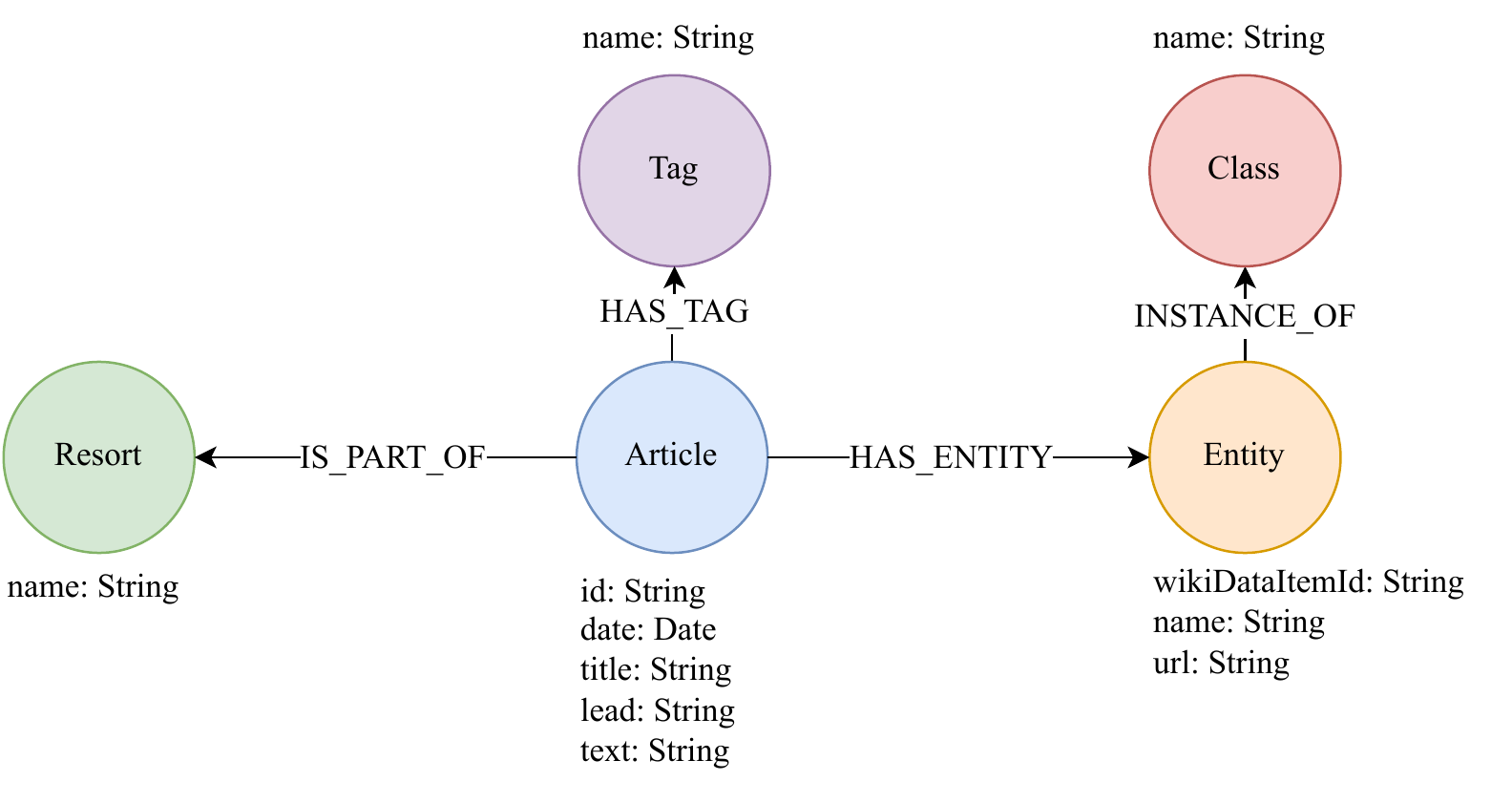}
  \caption{Data model of constructed news knowledge graph}
  \Description{Enter description}
\label{fig:data-model}
\end{figure}

The knowledge graph of our research prototype is maintained in a graph database, as previously described in Section~\ref{sec:entity-search}. For this purpose, we decided to use the graph database management system Neo4j \footnote{Neo4j: https://neo4j.com/}. In Figure \ref{fig:data-model}, the data model of the knowledge graph is illustrated in diagrammatic form. The knowledge graph is composed of distinct nodes, such as articles, resorts, tags, entities, and entity classes. Each article is associated with multiple properties, including a unique identifier, a creation date, a title, an opening paragraph, and an article text. In addition, articles can be labeled with so-called tags, which each have a unique name as a property. As such, news articles can have several relationships with tags. An article is part of one resort, which, similar to tags, only has a unique name as a property. Examples of resorts are the news categories economy or sports. The described nodes and relationships are constructed using semi-structured news data, which is automatically fetched every hour from the Tagesschau application programming interface (API) \footnote{Tagesschau API: https://tagesschau.api.bund.dev/}. The Tagesschau is the oldest and most famous news program in Germany. 

The graph database ensures that each node has a unique identifier through the use of constraints, as mentioned earlier. According to these restrictions, if a new node is supposed to be created with an already associated identifier, it will not be created. Instead, a relationship with the existing node is established. A news article can have multiple linked entities. To derive entities for a given article, we perform named entity recognition and entity linking. As a result, all entities have a unique identifier, the wikiDataItemId, a name, and a URL as properties. We make use of the Wikifier model \footnote{Wikifier: https://wikifier.org/}, which also provides an API endpoint. The model first extracts all named entities from the news article text. Then, they are linked to the target knowledge base Wikipedia. After the creation of the entity node in the knowledge graph, it is linked to the associated article. Besides, the wikiDataItemId and URL, which are set as the entity node’s properties, can be used to obtain additional information about an entity from Wikipedia. Similar to the other aforementioned nodes, entities with the same identifier are not generated twice since the wikiDataItemid is used as the unique identifier in the graph database. The knowledge graph offers the possibility to execute advanced queries containing multiple entities. Lastly, to enable even more intricate queries, all constructed entities are categorized into entity classes like city, company, or person.

\subsection{Conversational Search Agent}
\begin{figure}[h]
  \centering
  \includegraphics[width=0.99\linewidth]{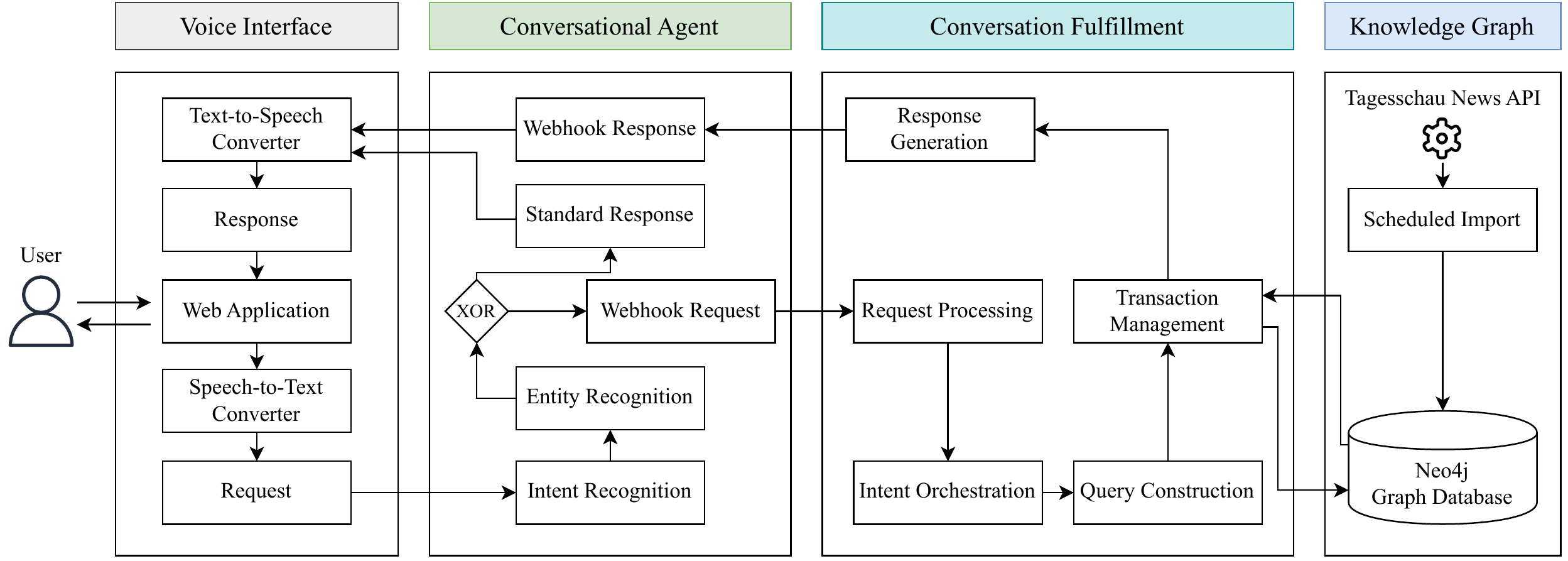}
  \caption{Architecture of conversational search agent}
  \Description{Enter description}
\label{fig:architecture}
\end{figure}

The search interface for the news knowledge graphs is developed in the form of a voice-based conversational agent. We used the natural language understanding platform Google Dialogflow \footnote{Google Dialogflow: https://cloud.google.com/dialogflow} because it provides a flexible API service to integrate conversational agents into websites, smart devices, or mobile apps. Although our final system will be an agent embedded in a smartwatch, for the initial prototype, we built a simple web application. Since the interaction happens via speech input and output, the website shows only one button for recording audio. The system architecture depicted in Figure~\ref{fig:architecture} supports the dialogue turns in the following manner. A user initiates the conversation with the agent by clicking the microphone button, whereupon the computer captures the user's utterance. Then, the input audio is transcribed with a speech-to-text model into text and sent to the Dialogflow agent. Next, the Dialogflow agent tries to predict the user's aim by using an intent recognition model that was fine-tuned with sample utterances. At this point, the agent also performs entity recognition to extract entities contained in the user’s query. Thereafter, based on the identified intent and entities, the agent decides whether a standard answer, such as a short salutation or default answer, is sent back to the user or a more complex response is necessary. If the latter is the case, a webhook request is sent to a Python-based web framework that extends the agent with a conversation fulfillment service. Based on the request’s intent and entity, the webhook service forwards it to the corresponding endpoint. After having received the request, we process it and match it to a Cypher query template, which is the query language for Neo4j. Once the query template is completed with parameters from the extracted entities, it is executed to retrieve the information from the knowledge graph. Finally, the query result from the graph database serves as input for the response generation. We construct the response by combining predefined textual elements and retrieved data items from the graph. We augment the text response with speech synthesis markup language (SSML) to annotate paragraphs, and sentences, as well as to insert pauses and emphasis, thereby enhancing the comprehensibility. In the last step, the response is sent back to the agent which applies a text-to-speech model to generate an audio file. This audio file is sent to the browser interface in is automatically played back to the user. 

The conversational agent was trained to recognize multiple intents encapsulating various user requests. For instance, there is a basic greeting intent that initiates the conversation after the agent briefly introduces its news search functionality. We separated the news search intents into three abstraction levels, similar to the question catalog in Table~\ref{tab:assistant-test}. First, users can ask for a general overview that returns three recent articles from three resorts. Second, it is possible to retrieve a list of resorts and then search for articles within one specific resort. Third, users can ask for news articles that are linked to specific entities. Additionally, there are intents recognizing navigation commands, such as selecting one of the suggestions or asking for other suggestions. In case of unexpected inputs that the agent cannot understand, a fallback intent is triggered. The latter hints to the user towards the help intent, which in turn, provides assistance and exemplary instructions on how to interact with the voice interface. Even though the agent has a finite set of intents, there is a virtually endless number of possible conversation paths that can occur due to daily changing news content in the graph.

\section{Preliminary Insights from Initial User Test}
\label{sec:user-study}
Before the planned evaluation study, in which we will deploy the conversational agent on a smartwatch and evaluate it with elderly people in a real-world setting, we conducted an initial test with a convenience sample of eight users. The idea behind this is to not only test the basic functionality but also identify technical and non-technical issues. The age range of participants was between 21 and 54, with an average age of 33.6, and the gender split was 37.5\% female and 62.5\% male. We did not give concrete instructions on how to use the agent, as we wanted to assess its self-explanatory nature. After the participants interacted with the agent through a web application, we asked them to report their experience in a free-form text. Overall, the test users liked the innovative concept of voice-based exploratory news search and praised the system's ease of use. Nobody had problems understanding the core functions. The participants appreciated both the ability to get a short general overview and to ask about specific subjects of their interest. They quickly learnt how to use the entity-based search and inquired about news related to various entities, such as the bank Credit Suisse, the person Mark Zuckerberg, or the country of Ukraine.

While it was not problematic for test users to start a news-seeking dialogue, they pointed out a number of suggestions for improvement. Some participants experienced technical problems with certain browsers, which impacted the microphone usage and the system's stability. This, among other reasons, caused problems in which the agent could not understand all user utterances. Regarding the recommended article headlines, the test users preferred a numerical enumeration over a simple listing so that they could choose by simply saying ''first'' or ''second article''. Other users wanted to be able to select articles by mentioning only a keyword from the headline. Besides, a few participants expressed the desire to customize the voice output, for example, they wanted to change the voice style or reading speed according to their needs. The collected user feedback informs the next development iteration of the agent.

\section{Conclusion and Future Outlook}
\label{sec:conclusion}
As the technology of conversational agents continues to evolve, it has the potential to significantly improve the quality of life for people with severe health conditions and the elderly. In the context of assisted living, seeking information about the daily news is a major use case. When people age or develop illnesses, they may experience impairments related to cognition, mobility, or vision, which makes accessing the news more challenging. 
Conversational interfaces can be an effective technology to alleviate these problems through natural interaction with the information medium.

In this paper, we investigated the development of conversational agents that support news search and exploration. We conducted an experimental analysis of commercial voice assistants and identified several weaknesses, including inadequate response formats, a lack of navigation control, or the absence of topic-specific search. To address these shortcomings, we propose a voice-based conversational agent for assisted living that enables users to explore daily news using simple dialogues. Our system builds upon a knowledge graph database that represents news articles in a structure for entity-based search and navigation. However, our prototype system is subject to a few limitations. For instance, the system was built for German-speaking users, since it will be evaluated in cooperation with care facilities in Germany. At the moment, the news data comes from a single source, namely the Tagesschau. Moreover, our initial prototype has only basic search capabilities that will be continually extended and improved based on the first user studies. Our system has been designed from the ground up to be easily extendable. We intend to conduct a more extensive user evaluation in the near future and add additional search and recommendation features. Due to the flexibility of the system architecture, we can quickly adapt the system to changing requirements of the user group. For the next development iterations, it is also planned to consider the integration of large language models and connect more information sources.

\begin{acks}
This work has been supported by funds from the Bavarian Ministry of Economic Affairs, Regional Development and Energy (StMWi) as part of the program “Bayerisches Verbundförderprogramm (BayVFP) – Förderlinie Digitalisierung – Informations- und Kommunikationstechnologie”. Kristiina Jokinen acknowledges the support of Project JPNP20006 commissioned by the New Energy and Industrial Technology Development Organization (NEDO), Japan.
\end{acks}

\bibliographystyle{ACM-Reference-Format}
\bibliography{sample-base}


\end{document}